\documentclass{llncs}

\usepackage{graphicx}
\usepackage{subcaption}
\usepackage{amsmath}
\usepackage{amssymb}
\usepackage{epsfig}
\usepackage{booktabs}

\usepackage{url}
\urldef{\mailsa}\path|{gsn2009,camilalaranjeira,viniciusbraz30}@ufmg.br, anisio@decom.cefetmg.br, erickson@dcc.ufmg.br|

\title{A Robust Indoor Scene Recognition Method based on Sparse Representation}

\author{Guilherme Nascimento%
\thanks{This work is supported by grants from Vale Institute of Technology, CNPq, CAPES, FAPEMIG and
CEFETMG; CNPq under Proc.  456166/2014-9 and 431458/2016-2; and FAPEMIG under Procs. APQ-00783-14 and APQ-03445-16, and FAPEMIG-PRONEX-MASWeb, Models, Algorithms and Systems for the Web APQ-01400-14.}
\and Camila Laranjeira \and Vinicius Braz \and \\ Anisio Lacerda%
\and Erickson R. Nascimento}

\institute{Universidade Federal de Minas Gerais (UFMG), Brazil\\
\mailsa\\}

\begin{document}

\maketitle

\begin{abstract}

In this paper, we present a robust method for scene recognition, which leverages Convolutional Neural Networks (CNNs) features and Sparse Coding setting by creating a new representation of indoor scenes. Although CNNs highly benefited the fields of computer vision and pattern recognition, convolutional layers adjust weights on a global-approach, which might lead to losing important local details such as objects and small structures.
Our proposed scene representation relies on both: global features that mostly refers to environment's structure, and local features that are sparsely combined to capture characteristics of common objects of a given scene.
This new representation is based on fragments of the scene and leverages features extracted by CNNs.
%
The experimental evaluation shows that the resulting representation outperforms previous scene recognition methods on Scene15 and MIT67 datasets, and performs competitively on SUN397, while being highly robust to perturbations in the input image such as noise and occlusion.
\end{abstract} 


\section{Introduction}

Scene recognition is one of the most challenging tasks on image classification, since the characterization of a scene is performed by using visual features of the objects that compose it and its spatial layout. 
In other words, a scene is the result of objects compositions. A given environment is more likely to be classified as a bathroom when equipped with a bath and a shower.
This is especially true when considering indoor images, which are the focus of this work.
The rationale is that besides the constituent objects, the image of a room (i.e., an indoor scene) is similar to every scene and it is hard to distinguish among them.

In this paper, we propose a novel method, which leverages features extracted by Convolutional Neural Networks (CNNs) using a sparse coding technique to represent discriminative regions of the scene. 
Specifically, we propose to combine global features from CNNs and local sparse feature vectors mixed with max spatial pooling. We built an over-complete dictionary whose base vectors are feature vectors extracted from fragments of a scene. This approach makes our image representation less sensitive to noise, occlusion and local spatial translations of regions and provides discriminative vectors with global and local features.

The main contributions of this paper are: 
i) a robust method that simultaneously leverages global and local features to the scene recognition task, and
ii) a thorough set of experiments to validate our requirements and the proposed scene recognition method. 
We evaluate our method on three different datasets (Scene15~\cite{FeiFei05cvpr}, MIT67~\cite{Quattoni09cvpr} and SUN397~\cite{Xiao2010cvpr}) comparing it to previously proposed methods in the literature, and perform a robustness test against the work of Herranz et al.~\cite{Herranz16cvpr}. The experimental results show that our method outperforms the current state of the art on Scene15 and MIT67, while performing competitively on SUN397. Additionally, when subjected to image perturbations (i.e., noise and occlusion), our proposal outperforms Herranz et al.~\cite{Herranz16cvpr} on all three datasets, surpassing it by a large margin on the most challenging one, SUN397.

\paragraph*{Related Work.}
\label{sec:rwork}

Most solutions proposed in the past decade for scene recognition exploited handcrafted  local~\cite{Jaakkola98nips} and global descriptors~\cite{Oliva2001ijvc}. 
Methods to combine these descriptors
and to build an image representation varied from 
Fisher Vectors~\cite{sanchez2013image} to Sparse Representation~\cite{oliveira2014tip}.
%
%
%
Sparse representation methods reached a great performance on image recognition, becoming popular in the last few years. 
Yang et al.~\cite{Yang09cvpr} and Gao et al.~\cite{Gao10cvpr} encoded SIFT~\cite{Lowe2004ijv} descriptors into a single sparse feature vector with the max-pooling method, achieving the best results on the Caltech-101 and the Scene-15 datasets.



In recent years, methods based on CNNs achieved state of the art on the task of scene recognition.
In Zhou et al.~\cite{Zhou14nips}, the authors introduce the Places dataset and present a CNN method to learn deep features for scene recognition.
The authors reached an accuracy of $70.8\%$ on MIT67. 
Another line of investigation is to combine features from CNNs. Dixit et al.~\cite{Dixit15cvpr} use a network to extract posterior probability vectors from local image patches and combine them using a Gaussian mixture model.
Features extracted from networks trained on ImageNet~\cite{Russakovsky15ijvc} and Places~\cite{Zhou14nips} are combined, achieving $79\%$ accuracy on MIT67.

Herranz et al.~\cite{Herranz16cvpr} analyzed the impact of object features in different scales in combination with global holistic features provided by scene features. According to the authors, the aggregation of local features brings robustness to semantic information in scene categorization. Herranz et al.~\cite{Herranz16cvpr} held the state of the art before Wang et al.~\cite{wang2017weakly} propose the architecture called PatchNet. This architecture models local representations from scene patches and aggregates them into a global representation using a newly semantic encoding method called VSAD.

Differently from previous works, our approach is flexible to being used with any feature representation, not being restricted to a single dataset or network architecture. This is a key advantage, since the use of different sources of features (e.g., ImageNet for object features and Places for structure) cannot be easily handled by training a single CNN or an autoencoder representation.
Additionally, our method is attractive because it presents a small number of hyperparameters (the sparsity and dictionary size), which makes it much easier to train. Typical CNN or autoencoder approaches require selecting the batch size, learning rate and momentum, architecture, optimization algorithm, activation functions, the number of neurons in the hidden layer and dropout regularization.





\section{Methodology}
\label{sec:methodology}

\begin{figure*}[t!]
    \centering
    \includegraphics[width=0.85\textwidth]{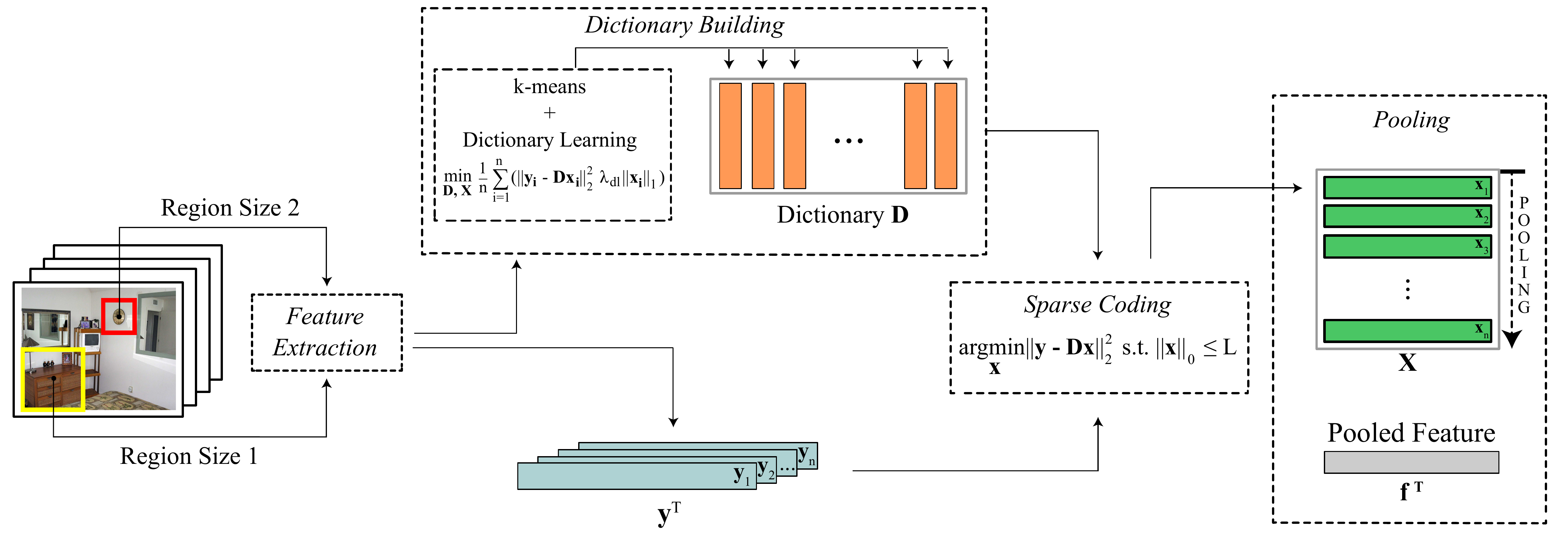} 
    \caption{Overview of the sparse representation pipeline. For each type of local feature, a single dictionary is built and optimized for both scales, and later used to encode the sparse vectors in $\mathbf{X}$. The final feature $\mathbf{f^T}$ of a given scale is computed by pooling all sparse representations calculated on that scale.}
    \label{fig:pooling_methodology}
\end{figure*}

Our methodology is composed of four steps, namely: i) feature extraction and dictionary building; ii) feature sparse coding; iii) pooling, and iv) concatenation. The final feature vector feeds a Linear SVM used to classify the image.  The three first steps are illustrated in Fig.~\ref{fig:pooling_methodology}.

The overall idea of our approach is to extract features from two semantic levels of the image. Firstly, a global representation is computed from the entire image, encoding the structural features of the scene. Then, we move a sliding window over the image with $\frac{I}{2}$ and $\frac{I}{4}$ window sizes, where $I$ represent the image dimensions, computing a feature vector for each fragment of the scene. These fragments contain local features that are potentially from objects or object's parts. 
The stride was fixed to half of the window dimensions in both scales. 
The rationale is that features extracted from smaller regions will convey information regarding the constituent objects of the scenes. Thus, by combining these features we can represent an indoor scene as a composition of objects.


\paragraph*{Feature Extraction and Dictionary Building.}
Although our method can be instantiated with any feature extractor, such as a bag-of-features model, we use  Convolutional Neural Networks to extract the features. 
We extract semantic features from the fc7 layer of VGGNet-16~\cite{Simonyan14corr}. For global information, the CNN was trained on Places~\cite{Zhou14nips}, while two types of local information were extracted: the same model trained on Places to acquire information regarding local structures, and a second model trained on ImageNet to encode object features. 

The feature extraction step provided a feature vector $\mathbf{y} \in \mathbb{R}^d$ for each fragment of the image. This process is performed for all samples, grouping the $\mathbf{y}$ vectors into $k$ clusters using the $k$-means algorithm. Hence, a dictionary for scale $i$ is represented by
$K_i = [\mathbf{v}_{1,i}, \ldots, \mathbf{v}_{k,i} ] \in \mathbb{R}^{d \times k}$,
where $\mathbf{v}_{j,i} \in \mathbb{R}^d$ represents the $j$th cluster in scale $i$.
We define the matrix $\mathbf{D_0}$ as the concatenation of multiple $c$ scales matrices $\{K_i|1 \leq i \leq c\}$: 
$\mathbf{D_0}= [K_1, K_2, \ldots, K_c]$.

It is worth noting that, since we are considering a set of over-complete basis, we need $\cup_{i=0}^c ~K_i \gg d$, where $K_i$ is the number of scale matrices in the dictionary and $d$ is the dimension of each feature vector. 

We concatenate the dictionaries built on different patch sizes into a single dictionary, respecting the nature of the feature. This leaves us with two dictionaries, one built from features extracted using the model trained on Places, and the other using the model trained on ImageNet.
The idea is improving the probability of finding a match in different scales, which offers a better discrimination and repeatability power. 
Once we have built the initial dictionary $\mathbf{D_0}$, we adjust the dictionary to matrix $\mathbf{D}$ by solving
\begin{equation} \label{eq:dictionary_learning}
\underset{\mathbf{\mathbf{D}, \mathbf{X}}}{\operatorname{min}} \quad \frac{1}{n} \sum_{i=1}^n \left(
\lVert \mathbf{y_i} - \mathbf{D} \mathbf{x_i} \rVert_2^2 \quad \lambda_{dl} \lVert \mathbf{x_i} \rVert_1 \right),
\end{equation}
\noindent where $\lVert \cdot \rVert_2$ is the $L2$-norm, $\lVert \cdot \rVert_1$ is the $L1$-norm, $\lambda_{dl}$ is a regularization parameter for the dictionary learning and the vector $\mathbf{x_i}$ is the $ith$ column of matrix $\mathbf{X}$. We applied a dictionary learning algorithm~\cite{Mairal09icml}, which solves Eq.~\ref{eq:dictionary_learning} alternating between $\mathbf{D}$ and $\mathbf{X}$ as variables, i.e., it minimizes over $\mathbf{D}$ while keeping $\mathbf{X}$ fixed. The dictionary $\mathbf{D_0}$ is used to start the optimization process.

\paragraph*{Sparse Coding.}

Despite the large number of possible objects that can compose a scene, each class of indoor environments has a small number of characteristic types of objects. 
The dictionary $\mathbf{D}$ provides a linear representation of each image fragment using a small number of basis functions. Thus, we find the sparse vector $\mathbf{x}$ by modeling the composition of a scene as a sparse coding problem.

Considering the new domain of representation defined by the dictionary $\mathbf{D}$, the representation of a feature vector $\mathbf{y}_i$ extracted from a sample of indoor class $i$, can be rewritten 
as $\mathbf{y}_i=\mathbf{D} \mathbf{x}_i$, where $\mathbf{x}_i$ is a vector whose entries are zero except those associated with the fragments of class $i$.


%


We use as a basis for the vector representation the columns $\mathbf{D}_i$ of a dictionary that is mixed with weights $\mathbf{x}$ to infer the vector $\mathbf{y}$, which is the vector that reconstructs the input $\mathbf{y}$ with the smallest error. Each column $\mathbf{D}_i$ is a descriptor representing an image fragment of the scene. We use a sparsity regularization term to select a small number of columns in the dictionary $\mathbf{D}$.

Let $\mathbf{y} \in \mathbb{R}^d$ be a descriptor extracted from an image patch and $\mathbf{D} \in \mathbb{R}^{d \times kc}$ ($d \ll kc$). The coding can be modeled as the following optimization problem:
\begin{equation} \label{eq:sparse_coding}
\mathbf{x^*} = \underset{\mathbf{x}}{\operatorname{argmin}}
\lVert \mathbf{y} - \mathbf{D} \mathbf{x} \rVert_2^2 \quad s.t. \quad \lVert \mathbf{x} \rVert_0 \leq L,
\end{equation}
\noindent where $\lVert \cdot \rVert_0$ is the $L0$-norm, indicating the number of nonzero values, and $L$ controls the sparsity
of vector $\mathbf{x}$. 
The final vector $\mathbf{x^*} \in \mathbb{R}^{kc}$  is the set of basis weights that represents the descriptor $\mathbf{y}$ as a linear and sparse combination of fragments of scenes.

Although the minimization problem of Eq. \ref{eq:sparse_coding} is NP-hard, greedy approaches, such as Orthogonal Matching Pursuit (OMP)~\cite{OMP2007TIT} or $L1$ norm relaxation, also know as LASSO~\cite{LASSO1994JRSS}, can be used to effectively solve it. 
In this paper, we use OMP because it achieved the best results on our tests.


%
%


The pooling process refers to the final step of the diagram presented in Fig.~\ref{fig:pooling_methodology}. To create the vector encoding the scene features, we compute 
the final feature vector $\mathbf{f} \in \mathbb{R}^{kc}$ by a pooling function $\mathbf{f}  = \mathcal{F}(\mathbf{X})$, where $\mathcal{F}$ is defined on each column of~$\mathbf{X}\in \mathbb{R}^{m \times kc}$. The rows of matrix $\mathbf{X}$ represent the sparse vectors of each feature vector extracted from $m$ sliding-windows.
We create $kc$-dimensional vectors according to the maximum pooling function, since according to Yang et al.~\cite{Yang09cvpr}, it gives the best results for sparse coded local features.


These steps are executed for both scales, using both the model trained on Places and ImageNet. At the end, five features vectors are concatenated into one: the global features as originally output by the CNN, and four local features.



\section{Experiments}
\label{sec:tests}




We evaluated our approach on the standard datasets for scene recognition benchmark: Scene15, MIT67 and SUN397. The specific attributes of each dataset allowed us to analyze different properties of our method. We trained the models on Places for global and local descriptors, and ImageNet for local descriptors, with VGGNet-16 which has shown the lowest classification error~\cite{Simonyan14corr}. When considering features from the VGGNet-16, we used PCA to reduce the descriptions from $4,096$ to $1,000$ on the second and third scales.

To compose the dictionary, we set $2,175$ words for Scene15, $3,886$ words for MIT67 and $6,907$ words for SUN397, according to the number of regions extracted from both scales and the number of classes of each dataset. The regularization factor $\lambda_{dl}$ for the dictionary learning (Eq.~\ref{eq:dictionary_learning}) is set to $0.1$. 
We set the same sparsity controller value $L$  (Eq.~\ref{eq:sparse_coding}) for all configurations when executing OMP. It was set to $0.03\times D_c$ non-zeros, where $D_c$ is the number of columns of the dictionary. 
All vectors were $L2$ normalized after each step and after concatenation. We used a Linear Support Vector Machine to perform the classification.


\paragraph*{Robustness evaluation.}

\begin{figure}[t!]
    \centering
    \begin{subfigure}[t]{0.15\textwidth}
        \centering
        \includegraphics[height=0.64in]{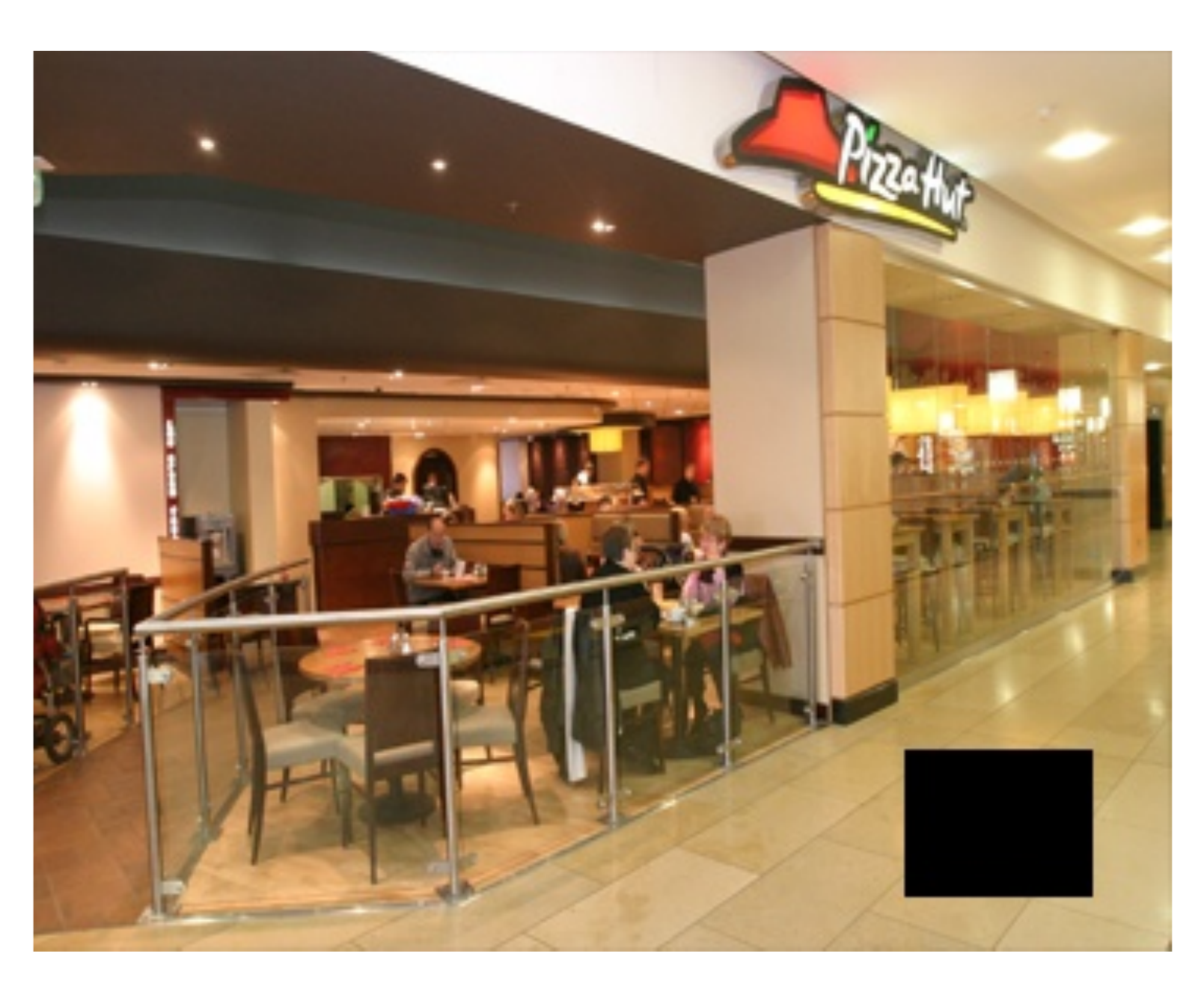}
    \end{subfigure}%
    ~
    \begin{subfigure}[t]{0.15\textwidth}
        \centering
        \includegraphics[height=0.64in]{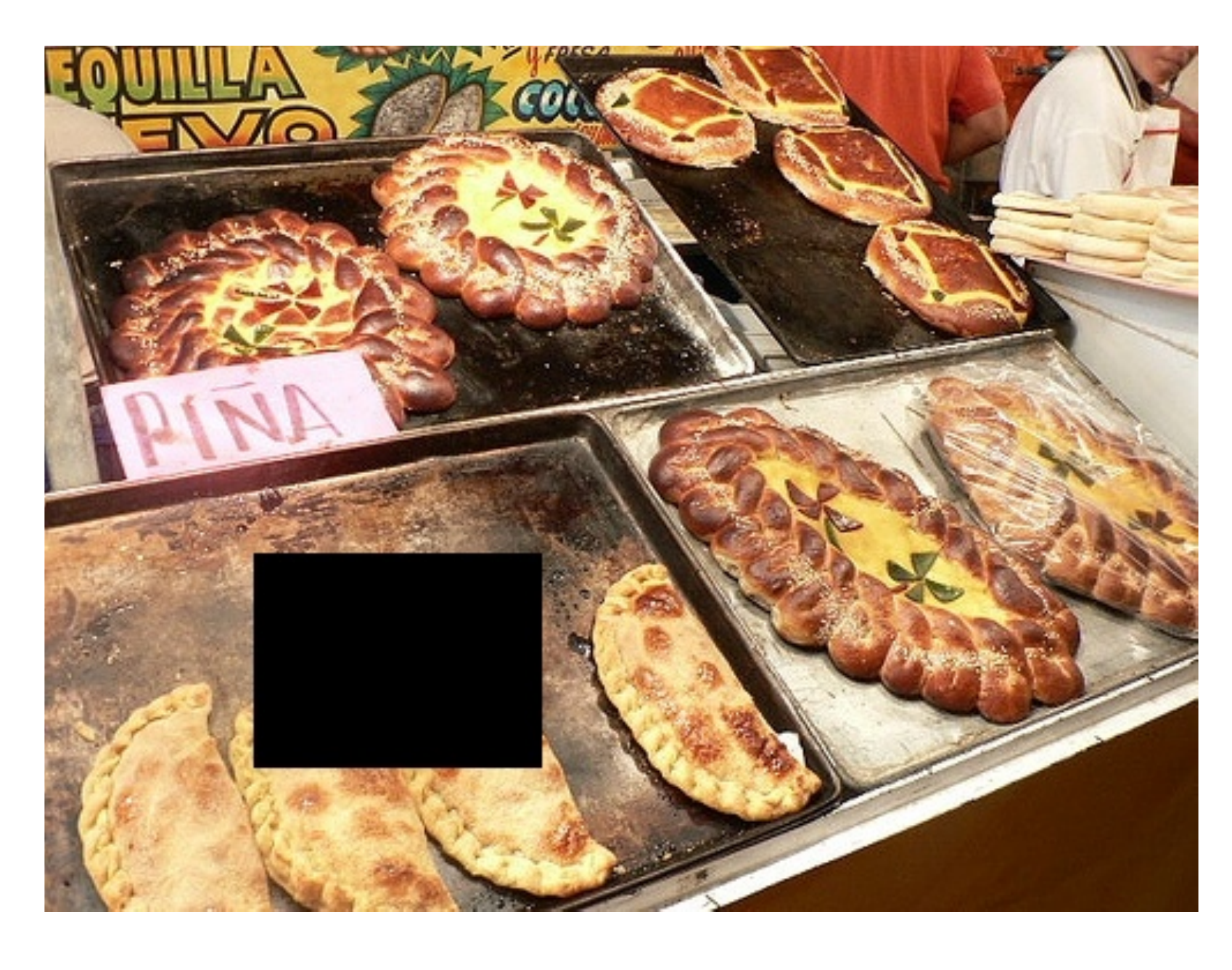}
    \end{subfigure}
    ~
    \begin{subfigure}[t]{0.15\textwidth}
        \centering
        \includegraphics[height=0.64in]{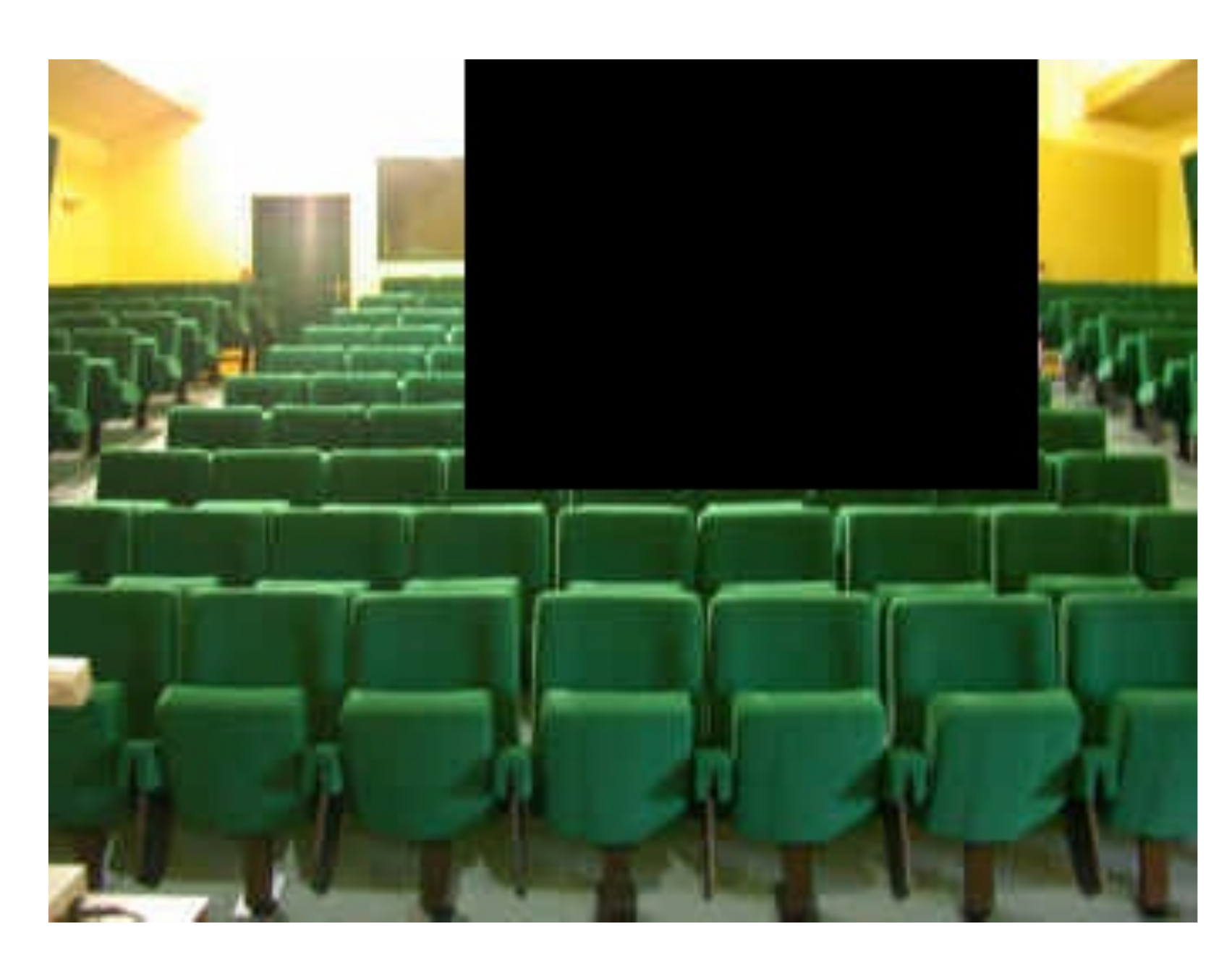}
    \end{subfigure}%
    ~
    \begin{subfigure}[t]{0.15\textwidth}
        \centering
        \includegraphics[height=0.64in]{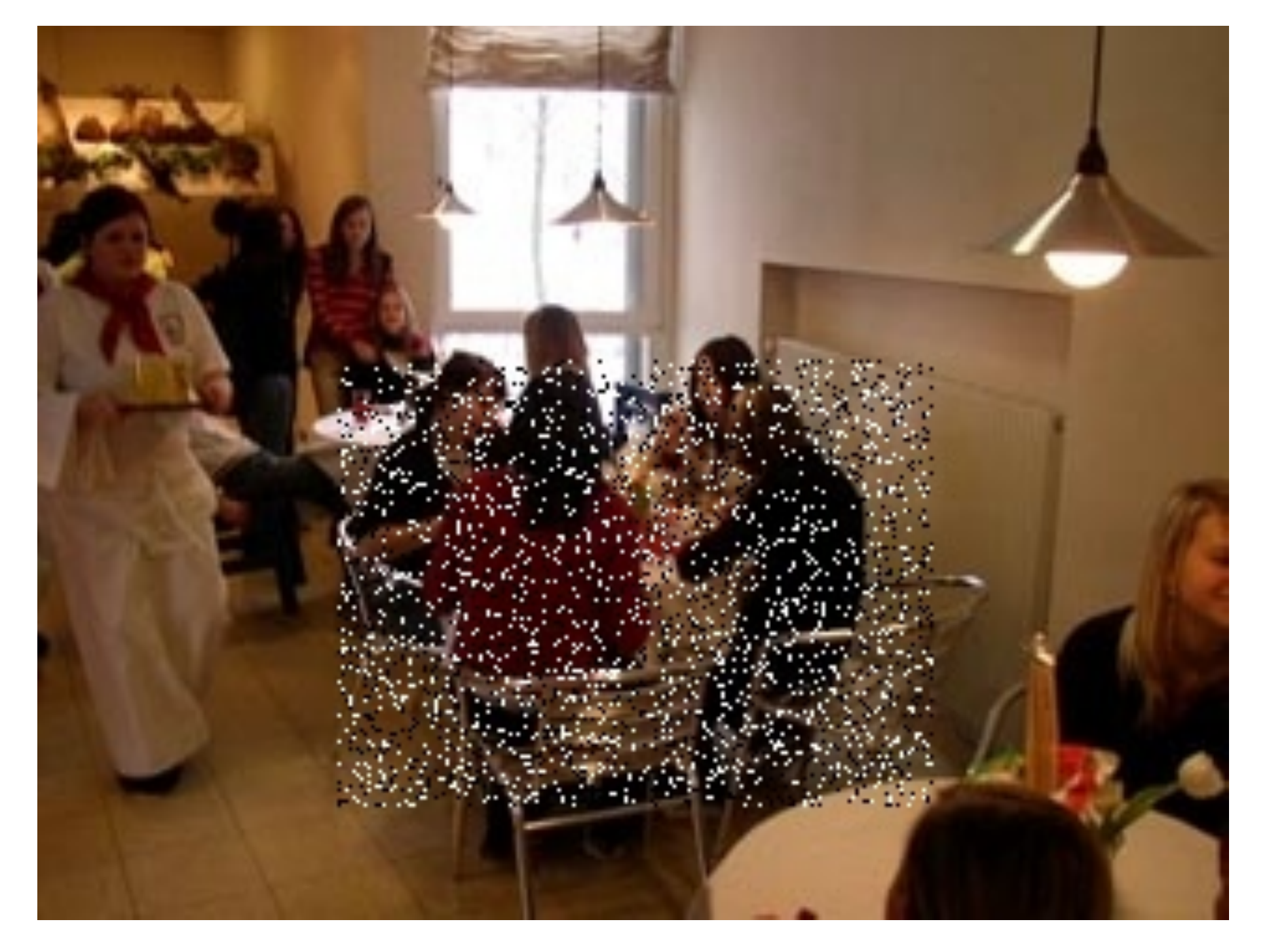}
    \end{subfigure}
    ~
    \begin{subfigure}[t]{0.15\textwidth}
        \centering
        \includegraphics[height=0.64in]{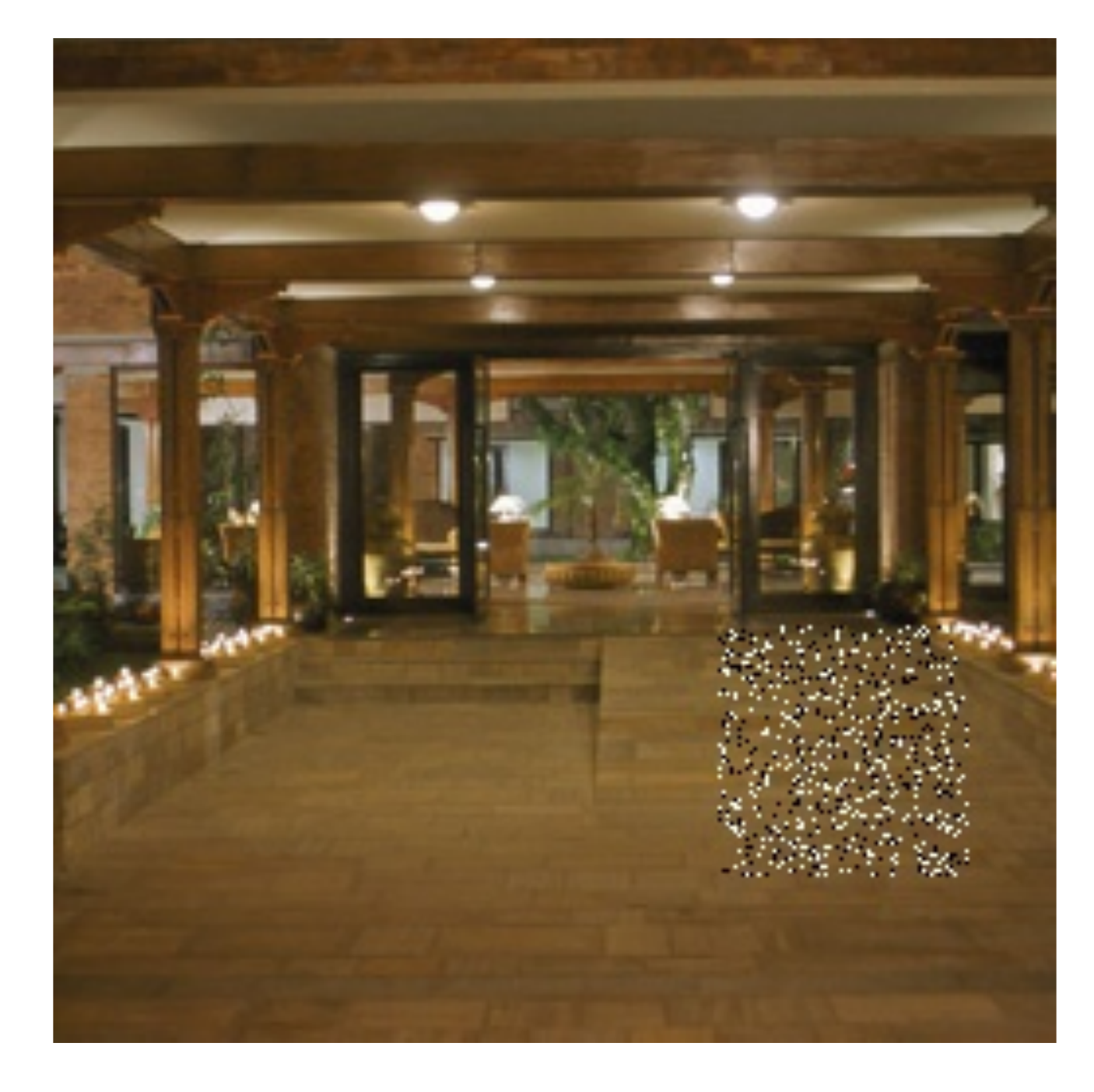}
    \end{subfigure}%
    ~
    \begin{subfigure}[t]{0.15\textwidth}
        \centering
        \includegraphics[height=0.64in]{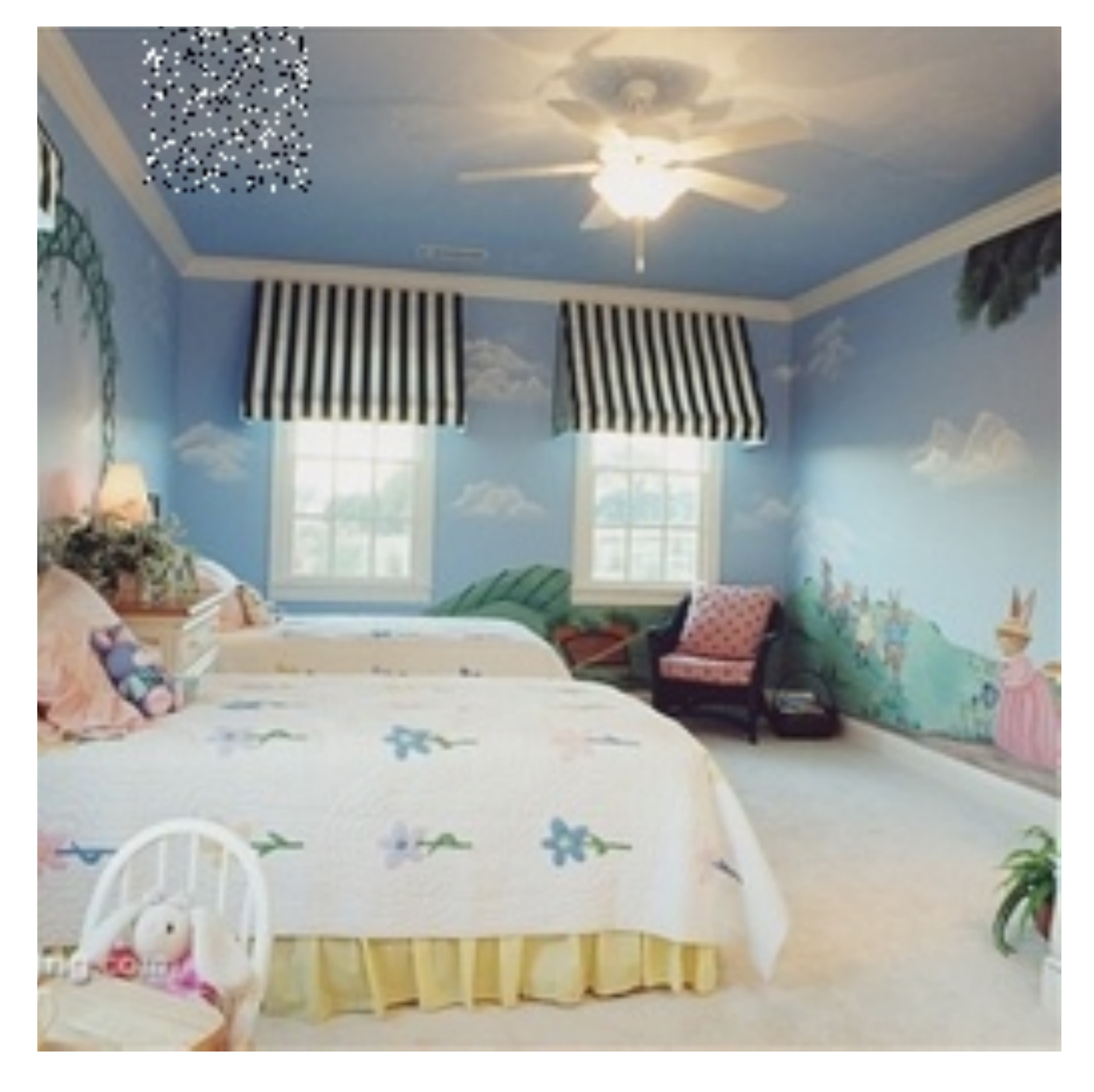}
    \end{subfigure}
    \caption{Occlusion and Noise examples. The location of windows was randomly computed, and experiments were performed for different sizes of windows.}
    \label{fig:ocnoise}
\end{figure}

We verified the descriptor robustness against occlusion and noise. For this purpose, we automatically generated squares randomly positioned, to simulate occlusions (black squares) and noise (granular squares), as illustrated in Fig. \ref{fig:ocnoise}. Experiments were performed for different sizes of windows $\frac{\mathbf{W}}{n}$, where $\mathbf{W}$ represents the dimensions of the image, with varying $n$ values.

Table \ref{tab:occnoise} presents the results for occlusion and noise robustness tested against Herranz et al.~\cite{Herranz16cvpr}. Our methodology performs better in all cases, indicating that our method is less sensitive to perturbations on the image. This behaviour is even more evident for SUN397, by far the most challenging of all three datasets, where the Herranz et al.'s methodology was inferior, on average $25.29\%$ for occlusion, and $23.05\%$ when noise was added. 

\begin{table}[t!]
\centering
\caption{ Accuracy performance comparison for Occlusion and Noise scenarios. Our approach shows higher accuracy performance. It provides a better feature selection for scene patches instead of just max-pooling between CNN features.}
\label{tab:occnoise}

\resizebox{\textwidth}{!}{%

\begin{tabular}{@{}llrrrrrrrrrrrrrr@{}}
\toprule 
 & & \multicolumn{4}{c}{Scene15}  & \phantom{abc} & \multicolumn{4}{c}{MIT67} & \phantom{abc} & \multicolumn{4}{c}{SUN397}  \\ \cmidrule{3-6} \cmidrule{8-11} \cmidrule{13-16}
\multicolumn{1}{r}{Window size}    &  & W/10 & W/8 & W/6 & \multicolumn{1}{r}{W/4} && W/10 & W/8 & W/6 & \multicolumn{1}{r}{W/4} && W/10 & W/8 & W/6 & \multicolumn{1}{r}{W/4} \\ \midrule 

Occlusion  & Herranz et al. & $94.07$ & $93.87$ & $93.97$  & \multicolumn{1}{r}{$92.93$}  && $83.79$ & $84.01$  & $83.79$ & \multicolumn{1}{r}{$83.56$}  &&  $39.95$  & $40.05$  & $38.99$ & \multicolumn{1}{r}{$35.61$}      \\ 
 
 & {\bf Ours}        & $94.43$  & $94.50$  & $94.37$ & \multicolumn{1}{r}{$93.50$} && $84.61$ & $84.61$  & $84.69$
 & \multicolumn{1}{r}{$84.76$} && $64.64$ & $64.71$ & $64.01$ & \multicolumn{1}{r}{$62.39$} \\ 
 & {\bf Avg. gain} & \multicolumn{4}{c}{$\mathbf{0.49}$} && \multicolumn{4}{c}{$\mathbf{0.88}$} && \multicolumn{4}{c}{$\mathbf{25.29}$} \\ \midrule
 
Noise  & Herranz et al. & $94.50$ & $94.27$ & $93.97$ & \multicolumn{1}{r}{$92.93$} && $83.94$ & $84.01$ & $83.49$ & \multicolumn{1}{r}{$81.32$} && $40.79$ & $40.42$ & $34.29$ & \multicolumn{1}{r}{$30.88$}  \\ 

 & {\bf Ours}  & $95.30$ & $94.57$ & $94.40$ & \multicolumn{1}{r}{$93.50$}  && $84.24$  & $85.28$  & $84.69$ & \multicolumn{1}{r}{$82.74$}  && $65.20$  & $65.29$  & $57.25$ & \multicolumn{1}{r}{$50.83$}  \\  

 & {\bf Avg. gain} & \multicolumn{4}{c}{$\mathbf{0.53}$} && \multicolumn{4}{c}{$\mathbf{1.05}$} && \multicolumn{4}{c}{$\mathbf{23.05}$} \\
  
   \bottomrule 
\end{tabular}

}

\end{table}

\paragraph*{Discussion.}

Besides being highly robust to perturbations in the input image, one can clearly see in Table~\ref{tab:stateofartresults} that our methodology also leads the performance on Scene15 and MIT67, while performing competitively on SUN397. It is worth highlighting that our method surpasses human accuracy, which was measured for SUN397 at $68.5\%$~\cite{Xiao2010cvpr}. To evaluate our assumption that local information can greatly benefit the task of indoor scene classification, Table~\ref{tab:stateofartresults} also presents the average accuracy achieved by a model trained solely on features from VGGNet-16 trained on Places, which composes the global portion of our methodology. On all datasets, global features by themselves showed inferior performance relative to our representation, revealing the complementary nature of local features. 

\begin{figure*}[t!]
\centering
\includegraphics[width=1\textwidth]{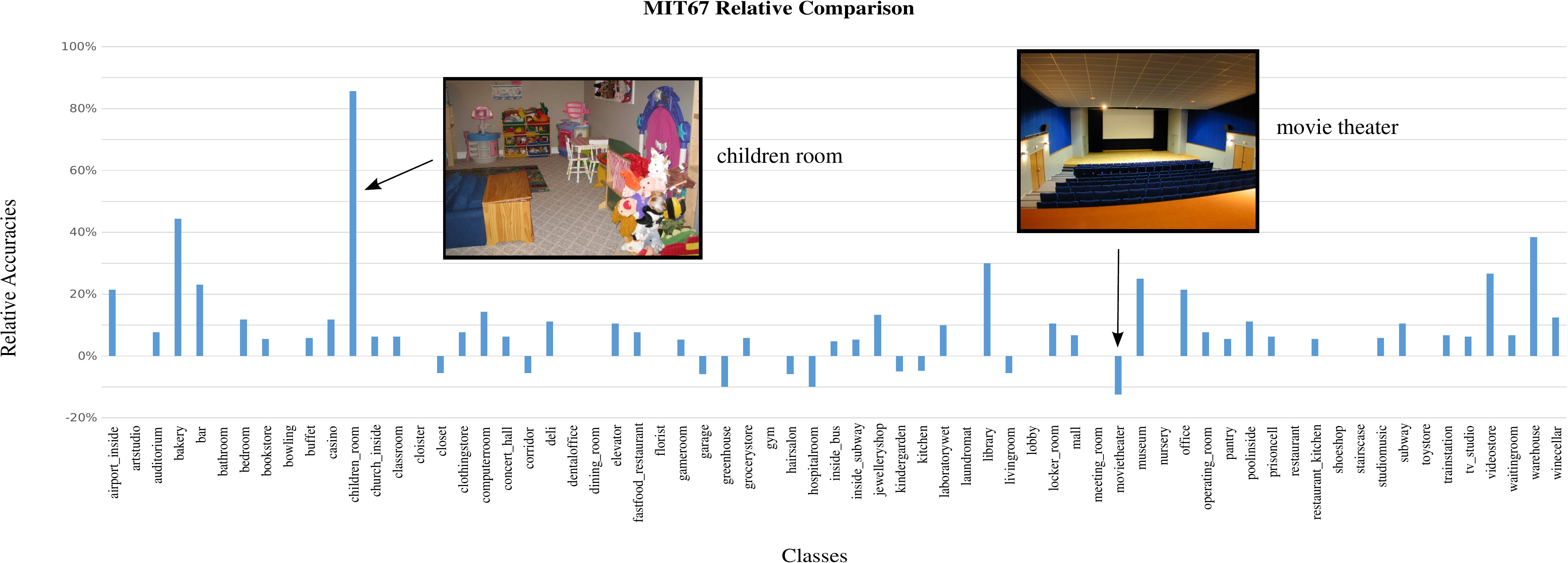} 
\caption{Relative accuracy between VGGNet-16 and our method on MIT67. In most cases, our method leads the performance.}
\label{fig:analysis}
\end{figure*}

We also performed a detailed performance assessment by comparing the accuracies among each indoor class on MIT67. The relative accuracies considering VGGNet-16 as a baseline are shown in Fig.~\ref{fig:analysis}. As we can see, for the classes that present a large number of object information (e.g., children room, emphasized in Fig. \ref{fig:analysis}) we have a significant increase in accuracy. 
We can draw the following two observations. 
First,  since the class of these environments strongly depends on the object configuration, the image fragments containing features of common objects can be represented by the sparse features on different scales. 
Second, for environments such as movie theater (Fig. \ref{fig:analysis}), the proposed approach is less effective, since such environments are strongly distinguished by their overall structure, and not necessarily by the constituent objects.

\begin{table}[t!]
\centering
\caption{Comparing average accuracy against other approaches.}
\label{tab:stateofartresults}
\small
\resizebox{\textwidth}{!}{\begin{tabular}{l@{\hskip 0.3in}lrrrrrrrrrrrrrr}
\toprule
Method & SPM~\cite{Lazebnik2006cvpr} & \phantom{a} & ScSPM~\cite{Yang09cvpr} & \phantom{a} & MOP-CNN~\cite{Gong14corr} & \phantom{a} & SFV~\cite{Dixit15cvpr} & \phantom{a} & VGGNet-16 & \phantom{a} & Herranz et al.~\cite{Herranz16cvpr} & \phantom{a} & VSAD~\cite{wang2017weakly} & \phantom{a} & \textbf{Ours} \\ \hline
Scene15 & $81.40\%$ && $80.28\%$ && -       &&   -     && $94.33\%$ && $95.18\%$ &&    -     && $\mathbf{95.73\%}$ \\
MIT67   &    -    &&    -    && $68.88\%$ && $79.00\%$ && $80.88\%$ && $86.04\%$ && $86.2\%$ && $\mathbf{87.22\%}$ \\
SUN397  &    -    &&    -    && $51.98\%$ && $61.72\%$ && $66.90\%$ && $70.17\%$ && $\mathbf{73.0\%}$ && $71.08\%$ \\
\bottomrule
\end{tabular}}
\end{table}

\section{Conclusion}
\label{sec:concfw}

We proposed a robust method that combines both global and local features to compose a high discriminative representation of indoor scenes. 
Our method improves the accuracy of CNN features by composing local features using Sparse-Coding and a max-pooling technique,
by creating an indoor scene representation based on an over-complete dictionary, which is built from several image fragments. 
Our sparse coding-based composition approach was capable of encoding local patterns and structures for indoor environments, and in cases of strong occlusion and noise, we lead the performance on scene recognition, which is explained by the advantages of sparse representations.
Our representation outperformed VGGNet-16 in all cases, reflecting the complementary nature of local features, and outperformed the current state of the art on Scene15 and MIT67, while being competitive on SUN397.



%

\bibliographystyle{splncs03}
\bibliography{bibliography}

\end{document}